\documentclass[runningheads]{llncs}

\usepackage{graphicx, caption, subcaption}
\usepackage{amssymb}
\usepackage{float}
\usepackage[utf8]{inputenc}
\usepackage[T1]{fontenc}
\usepackage{listings}
\usepackage{array, caption, tabularx, makecell, booktabs}%
\usepackage{multirow}
\usepackage{adjustbox}
\usepackage{bbm}
\usepackage{bbold}

\usepackage[frozencache, cachedir=.]{minted}
\usepackage[colorlinks = true, linkcolor = blue, urlcolor = blue, citecolor = black, anchorcolor = blue]{hyperref}

\begin{document}
\newcounter{foo}
\author{Jagriti Sikka\inst{1} \and
 Kushal Satya\inst{1} \and
 Yaman Kumar\inst{1} \and Shagun Uppal\inst{2} \and Rajiv Ratn Shah\inst{2} \and Roger Zimmermann\inst{3}}
 \institute{Adobe, Noida \and Midas Lab, IIIT Delhi \and School of Computing, National University of Singapore\\ \email{\{jsikka,satya,ykumar\}@adobe.com  \{shagun16088,rajivratn\}@iiitd.ac.in rogerz@comp.nus.edu.sg}}
\title{Learning based Methods for Code Runtime Complexity Prediction}
%
%
%
%
\maketitle              
\begin{abstract}
Predicting the runtime complexity of a programming code is an arduous task. In fact, even for humans, it requires a subtle analysis and comprehensive knowledge of algorithms to predict time complexity with high fidelity, given any code. As per Turing's Halting problem proof, estimating code complexity is mathematically impossible. Nevertheless, an approximate solution to such a task can help developers to get real-time feedback for the efficiency of their code. 
In this work, we model this problem as a machine learning task and check its feasibility with thorough analysis. Due to the lack of any open source dataset for this task, we propose our own annotated dataset \emph{CoRCoD: Code Runtime Complexity Dataset}\footnote{The complete dataset is available for use at \href{https://github.com/midas-research/corcod-dataset}{https://github.com/midas-research/corcod-dataset}.}, extracted from online judges. 
We establish baselines using two different approaches: feature engineering and code embeddings, to achieve state of the art results and compare their performances. 
Such solutions can be widely useful in potential applications like automatically grading coding assignments, IDE-integrated tools for static code analysis, and others.  

\end{abstract}
\section{Introduction}
Time Complexity computation is a crucial aspect in the study and design of well-structured and computationally efficient algorithms. It is a  measure of the performance of a solution for a given problem. As a popular mistaken consideration, it is not the execution time of a code. Execution time depends upon a number of factors such as the operating system, hardware, processors etc. Since execution time is machine dependent, it is not used as a standard measure to analyze the efficiency of algorithms. 
Formally, \emph{Time Complexity} quantifies the amount of time taken by an algorithm to process as a function of the input. For a given algorithm, we consider its worst case complexity, which reflects the maximum time required to process it, given an input. Time complexity is represented in \textbf{Big O} notation. $O(n)$ denotes the asymptotic upper bound of an algorithm as a function of the input size $n$. Typically, the complexity classes in Computer Science refer to P, NP classes of decision problems, however, for the entire length of this paper, complexity class refers to a category of time complexity. The commonly considered categories in computer science as well in our work are $O(1)$, $O(logn)$, $O(n)$ , $O(nlogn)$ and $O(n^2)$. 

In this work, we try to predict the time complexity of a solution, given the code. This can have widespread applications, especially in the field of education. It can be used in automatic evaluation of code submissions on different online judges. It can also aid in static analyses, informing developers how optimized their code is, enabling more efficient development of industry level solutions.

A number of ways have been proposed to estimate code complexity.  McCabe et al. \cite{mccabe} is one of the earliest works done in this area, which proposed \textbf{Cyclomatic Complexity}. It is defined as the quantitative measure of the number of linearly independent paths through a program’s source code computed using the control flow graph of the program. 
Cyclomatic Complexity of a code can be calculated through Equation \ref{Equation1}. 
\begin{equation}
C = E - N + M
\label{Equation1}
\end{equation}
where $E$ is the number of the edges, $N$ is the number of nodes and $M$ is the number of connected components.

Cyclomatic Complexity  quantitatively measures a program's logical strength based on existing decision paths in the source code. However, the number of independent paths does not represent how many times these paths were executed. Hence, it is not a robust measure for time complexity.


Bentley et al. \cite{master} proposed the \textbf{Master Theorem}. For the generic divide and conquer problem which divides a problem of input size $n$ into $a$ subproblems each of size $\frac{n}{b}$ ,and combine the result in $f(n)$  operations, Master theorem expresses the recurrence relation as Equation \ref{masters theorem}.
\begin{equation}
 T(n) = aT(\frac{n}{b}) + f(n)
 \label{masters theorem}
\end{equation}

However, the master theorem is limited to divide and conquer problems and has several constraints on the permissible values of $a$, $b$ and $f(n)$. 

Mathematically speaking, it is impossible to find a universal function to compute the time complexity of all programs. Rice's theorem and other works in this area \cite{haltingproof,ricetheorem} have established that the runtime complexity for problems in category P is undecidable i.e. it is impossible to formulate a mathematical function to calculate the complexity of any code with polynomial order complexity. 

Therefore, we need a Machine Learning based solution which can learn the internal structure of the code effectively. Recent research in the areas of machine learning and deep learning for programming codes provide several potential approaches which can be extended to solve this problem \cite{code2vec,graph2vec}. Also, several \emph{"Big Code"} datasets have been made available publicly. The Public Git Archive is a dataset of a large collection of Github repositories \cite{public_git_archive_dataset}, \cite{so_dataset} and \cite{StacQ} are datasets of Question-code pairs mined from Stack Overflow. However, to the best of our knowledge, at the time of writing this paper, there is no existing public dataset that, given the source code, gives runtime complexity of the source code. In our work, we have tried to address this problem by creating a Code Runtime Complexity Dataset \emph{(CoRCoD)} consisting of 932 code files belonging to 5 different classes of complexities, namely \emph{$O(1)$, $O(logn)$, $O(n)$, $O(nlogn)$} and \emph{$O(n^2)$} (see Table \ref{table:classwise_data}). 
\\

We aim to substantially explore and solve the problem of code runtime complexity prediction using machine learning with the following contributions:

\begin{itemize}
\item [$\bullet$] Releasing a novel annotated dataset of program codes with their runtime complexities. 

\item [$\bullet$] Proposing baselines of ML models with hand-engineered feature and study of how these features affect the computational efficiency of the codes. 

\item [$\bullet$] Proposing another baseline, the generation of code embeddings from Abstract Syntax Tree of source codes to perform classification.
\end{itemize}
Furthermore, we find that code embeddings have a comparable performance to hand-engineered features for classification using Support Vector Machines (SVMs). To the best of our knowledge, CoRCoD is the first public dataset for code runtime complexity, and this is the first work that uses Machine Learning for runtime complexity prediction. \\


The rest of this paper is structured as follows. In Section 3, we talk about dataset curation and its key characteristics.
We experiment using two different baselines on the dataset: classification using hand engineered features extracted from code and using graph based methods to extract the code embeddings via Abstract Syntax Tree of code.  Section 4 explains the details and key findings of these two approaches. In Section 5, we enumerate the results of our model and data ablation experiments performed on these two baselines.

\section{Related Work}
In recent years, there has been extensive research in the deep learning community on programming codes. 
Most of the research has been focused on two buckets, either on predicting some structure/attribute in the program or generating code snippets that are syntactically and/or semantically correct. In the latter case, Sun et al. \cite{Sun_Zhu_Mou_Xiong_Li_Zhang_2019} made use of the Abstract Syntax Tree (AST) of a program, and generated code by predicting the grammar rules using a CNN.

Variable/Method name prediction is a widely attempted problem, wherein Allamanis et al. \cite{pmlr-v48-allamanis16} used a convolutional neural network with attention technique to predict method names, Alon et al. \cite{ASTPaths-code2vec} suggested the use of AST paths to be used as context for generating \emph{code embeddings} and training classifiers on top of them. Yonai et al. \cite{MethodNamePred-CallGraphEmbedding} used call graphs to compute method embeddings and recommend names of existing methods with function similar to target function.

Another popular prediction problem is that of defect prediction, given a piece of code. Li et al. \cite{DefectPred} used Abstract Syntax Trees of programs in their CNN for feature generation which are then used for defect prediction. A major goal in all these approaches is to come up with a representation of the source program, which effectively captures the syntactic and semantic features of the program. Chen and Monperrus \cite{CodeEmbeddings-LiteratureStudy} performed a survey on word embedding techniques used on source codes. However, so far, there has been no such work for predicting time complexity of programs using code embeddings. We have established the same as one of our baselines using graph2vec \cite{graph2vec}.

Srikant and Aggarwal\cite{AspiringMinds} extract hand-engineered features from Control Flow and Data Dependency graphs of programs such as number of nested loops, number of instances of \emph{if} statements in a $loop$ etc. for automatic grading of programs. They then used the grading criteria, that correct test programs would have similar programming constructs/features as those in the correct hand-graded programs. We use the same idea of identifying key features as the other baseline, which are constructs that a human evaluator would look at, to compute complexity and use them to train the classification models. Though, unlike \cite{AspiringMinds}, our features are problem independent. Moreover, their dataset is not publicly available. 

\section{Dataset}

To construct our dataset, we collected source codes of different problems from Codeforces\footnote{https://codeforces.com}. Codeforces is a platform that regularly hosts programming contests. The large availability of contests having a wide variety of problems both in terms of data structures and algorithms as well as runtime complexity, made Codeforces a viable choice for our dataset.

\begin{table}[h]
    \begin{minipage}{.5\linewidth}
      \centering
            \begin{tabular}{|c|c|}
            \hline
            Complexity class & Number of samples\\
            \hline\hline
            $O(n)$ & 385 \\
            $O(n^2)$ & 200 \\
            $O(nlogn)$ & 150 \\
            $O(1)$ &  143 \\
            $O(logn)$ & 55 \\
            \hline\hline
            \end{tabular}
            \caption{Classwise data distribution}
            \label{table:classwise_data}

    \end{minipage}%
    \begin{minipage}{.5\linewidth}
      \centering
        \begin{tabular}{|c|c|}
            \hline
            \multicolumn{2}{|c|}{\bf Features from Code Samples}\\ \hline
            Number of methods & Number of breaks  \\ \hline
            Number of switches & Number of loops \\ \hline
            Priority queue present & Sort present \\ \hline
            Hash map present & Hash set present \\ \hline
            Nested loop depth & Recursion present \\ \hline
            Number of variables & Number of ifs \\ \hline
            Number of statements & Number of jumps \\ \hline
            \end{tabular}
            \caption{Extracted features} 
        \label{table:features}

    \end{minipage} 
\end{table}



For the purpose of construction of our dataset, we collected Java source codes from Codeforces. We used the Codeforces API to retrieve problem and contest information, and further used web scraping to download the solution source codes.

In order to ensure correctness of evaluated runtime complexity, the source codes selected should be devoid of issues such as compilation errors and segmentation faults. To meet this criterion, we filtered the source codes on the basis of their verdict and only selected the codes having verdicts \emph{Accepted} or \emph{Time limit exceeded} (TLE). For codes having TLE verdict, we ensured accuracy of solutions by only selecting codes that successfully passed at least four Test Cases. This criterion also allowed us to include multiple solutions for a single problem, different solutions having different runtime complexities. 
These codes were then manually annotated by a group of five experts, hailing from programming background each with a bachelor’s degree in Computer Science or related field. 
Each code was analyzed and annotated by two experts, in order to minimize the potential for error. Only the order of complexity was recorded, for example, a solution having two variable inputs, $n$ and $m$, and having a runtime complexity of $O(n*m)$ is labeled as $n\_square$ ($O({n}^2)$). 

Certain agreed upon rules were followed for the annotation process. The rationale relies on the underlying implementations of these data structures in Java. Following points list down the rules followed for annotation and the corresponding rationale:

\begin{itemize}
\item [$\bullet$] Sorting algorithm’s implementation in Java collections has worst case complexity $O(nlogn)$.

\item [$\bullet$] Insertion/retrieval in HashSet and HashMap is annotated to be $O(1)$, given $ n $ elements. 

\item [$\bullet$]  TreeSet and TreeMap are implemented as Red-Black trees and thus have $O(logn)$ complexity for insertion/retrieval.
\end{itemize}
\begin{listing}

\label{Algorithm1}
\begin{minted}[breaklines,frame=single]{java}

class noOfNestedLoops extends ASTVisitor {
    int current = 0;
    int max_depth = 0;
    @Override
    bool visit(WhileStatement node) {
        current += 1;
        max_depth = max(current, max_depth);
        return true;
    }
    @Override
    void endVisit(WhileStatement node){
        current -= 1;
    }
}
\end{minted}
\caption{Extracting number Of nested loops using ASTVisitor}\label{Algorithm1}
\end{listing}


We removed few classes with insufficient data points, and ended up with 932 source codes, 5 complexity classes, corresponding annotation and extracted features. We selected nearly 400 problems from 170 contests, picking an average of 3 problems per contest. For 120 of these problems, we collected 4-5 different solutions, with different complexities. 


\section{Solution Approach}
The classification model is trained using two approaches: one, extracting hand-engineered features from code using static analysis and two, learning a generic representation of codes in the form of code embeddings.
\subsection{Feature Engineering}

\begin{figure}[h]
    \centering
    \includegraphics[height=8.5cm, width=9.5cm]{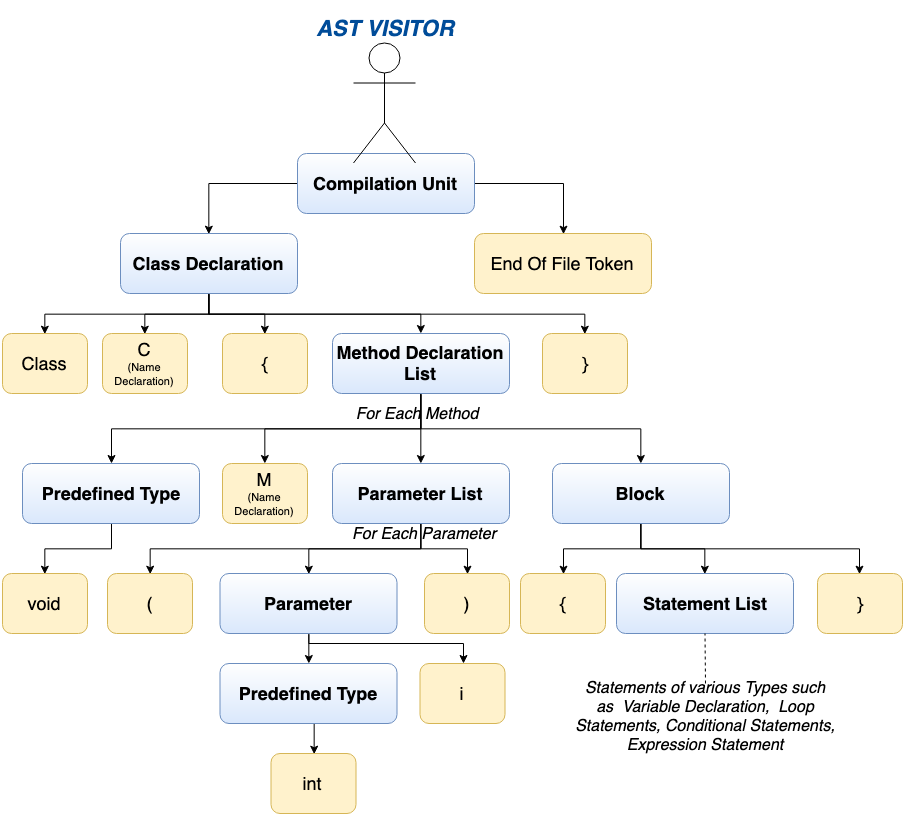}
    \caption{Extraction of features from code using AST Parser}
    \label{fig:AST}
\end{figure}

\paragraph{Feature Extraction.}
We identified key coding constructs and extracted 14 features (refer Table \ref{table:features}).
We extracted these features from the Abstract Syntax Tree (AST) of source codes. AST is a tree representation of syntax rules of a programming language. ASTs are used by compilers to check codes for accuracy. We used Eclipse JDT for feature extraction. A generic representation of AST as parsed by ASTParser in JDT is shown in Figure \ref{fig:AST}.

An ASTParser object creates the AST, and the ASTVisitor object \emph{"visits"} the nodes of the tree via \emph{visit} and \emph{endVisit} methods using Depth First Search.  
One of the features chosen was the maximum depth of nested loops. Code snippet in Listing ~\ref{Algorithm1} depicts how the value of depth of Nested Loops was calculated using ASTVisitor provided by JDT. Other features were calculated in a similar manner.



We observed that our code samples often have unused code like methods or class implementations never invoked from the main function. 
Removing such unused code manually from each code sample is tedious.
Instead, we used JDT plugins to identify the methods reachable from main function and used those methods for extracting listed features. The same technique was also used while creating the AST for the next baseline.

\begin{figure}[h]
    \begin{subfigure}[t]{0.5\textwidth}
      \includegraphics[width=\textwidth]{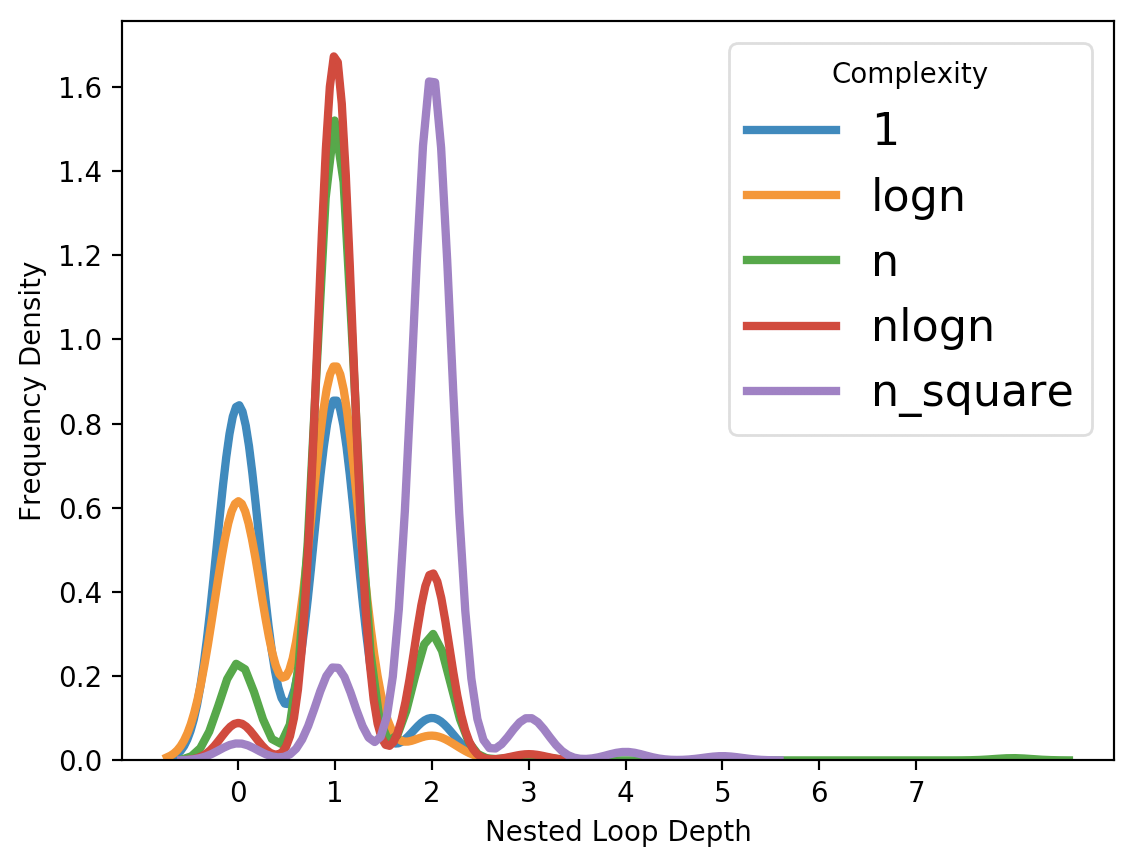}
      \caption{Depth of nested loop}
       \label{fig:loop}
    \end{subfigure}
    \begin{subfigure}[t]{0.5\textwidth}
      \includegraphics[width=\textwidth]{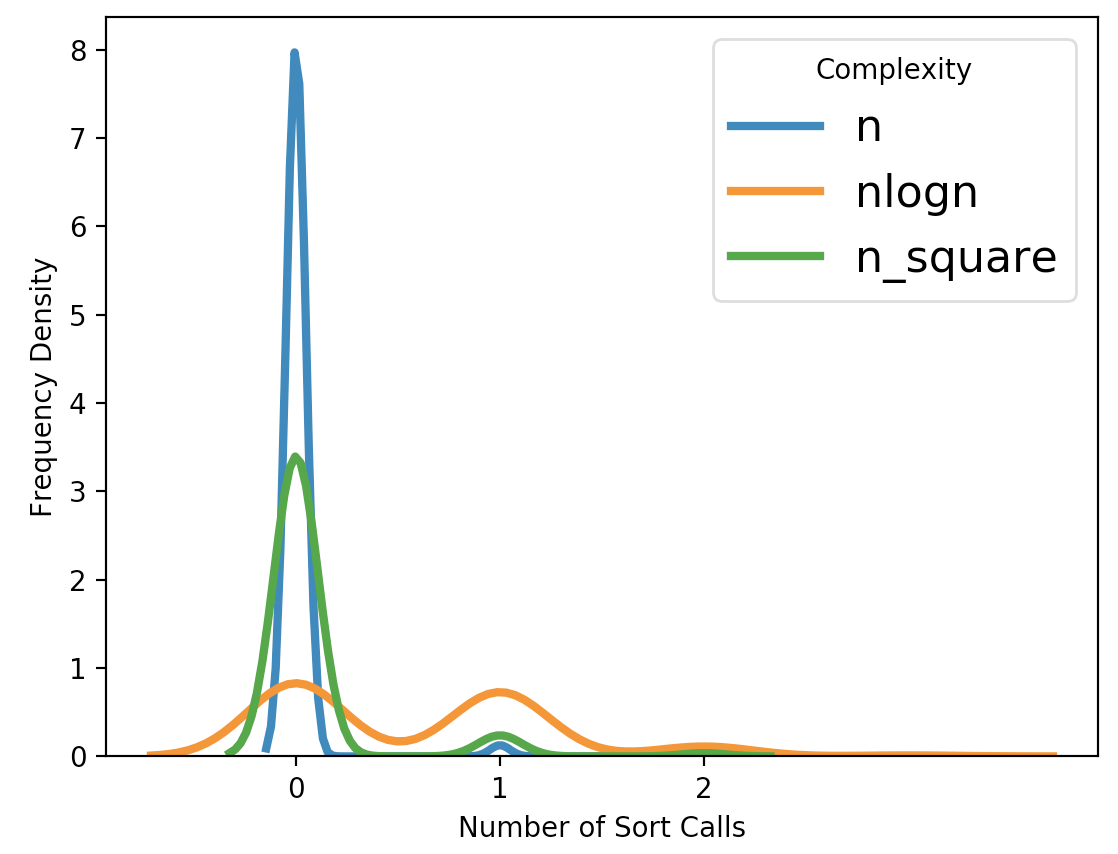}
      \caption{Number of sort calls}
      \label{fig:sorts}
    \end{subfigure}
    \caption{Density plot for the different features}
    \label{fig:density plot}
  \end{figure}
 

Figure \ref{fig:density plot} represents the  density distribution of features across different classes. For nested loops, $n\_square$ has peak at depth $2$ as expected; similarly $n$ and $nlogn$ have peak at depth $1$ loop depth (see Figure \ref{fig:density plot}(\subref{fig:loop})). For number of sort calls, $n$ and $n\_square$ have higher peaks at $0$, indicating peak in the absence of sort whereas $nlogn$ peaks both at $0$ and $1$ (see Figure \ref{fig:density plot}(\subref{fig:sorts})). Upon qualitative analysis, we found that a large number of solutions were performing binary search within a \emph{for} loop, which explains the peak at $0$, and others were simply using sort, hence the peak at $1$. This confirms our intuition that sort calls and nested loops are important parameters in complexity computation. 

\subsection{Code Embeddings}

The Abstract Syntax Tree of a program captures comprehensive 
information regarding a program's structure, syntactic and semantic relationships between variables and methods. An effective method to incorporate this information is to compute code embeddings from the program's AST. We use graph2vec
, a neural embedding framework \cite{graph2vec}, to compute embeddings. Graph2vec automatically generates task agnostic embeddings, and does not require a large corpus of data, making it apt for our problem.We used the graph2vec implementation from \cite{graph2vec_implementation} to compute code embeddings.

Graph2vec is analogous to doc2vec\cite{doc2vec} which predicts a document embedding given the sequence of words in it. The goal of graph2vec is, given a set of graphs $\mathbb{G} = \{G_1, G_2, ... G_n\}$, learn a $\delta$ dimensional embedding vector for each graph. Here, each graph $G$ is represented as $(N,E,\lambda)$ where $N$ are the nodes of the graph, E the edges and $\lambda$ represents a function $n \rightarrow l$  which assigns a unique label from alphabet $l$ to every node $n \in N$.  To achieve the same, graph2vec extracts nonlinear substructures, more specifically, rooted subgraphs from each graph which are analogical to words in doc2vec. It uses skipgram model for learning graph embeddings which correspond to code embeddings in our scenario. The model works by considering a subgraph $s_j \in c(g_i)$ to be occurring in the context of graph $g_i$ and tries to maximize the log likelihood in Equation \ref{loglikelihood}:

\begin{equation}
    \sum_{j=1}^{D}  log \; Pr({s_j}|{g_i})
    \label{loglikelihood}
\end{equation}
where $c(g_i)$ gives all subgraphs of a graph $g_i$ and $D$ is the total number of subgraphs in the entire graph corpus.

We extracted AST from all codes using the JDT plugins. Each node in AST has two attributes: a Node Type and an optional Node Value. For e.g., a MethodDeclaration Type node will have the declared function name as the node value. Graph2vec expects each node to have a single label. To get a single label, we followed two different representations:

\begin{enumerate}
    \item Concatenating Node Type and Node Value.
    \item Choosing selectively for each type of node whether to include node type or node value. 
    For instance, every identifier node has a SimpleName node as its child. For all such nodes, only node value i.e. identifier name was considered as the label.
\end{enumerate}

For both the AST representations, we used graph2vec to generate 1024-dimensional code embeddings. These embeddings are further used to train SVM based classification model and several experiments are performed as discussed in the next section. 
\section{Experiments and Results}
\subsection{Feature Engineering}
Deep Learning(\emph{DL}) algorithms tend to improve their performance with the amount of data available unlike classical machine learning algorithms. With lesser amount of data and correctly hand engineered features, Machine Learning(\emph{ML}) methods outperform many \emph{DL} models. Moreover, the former are computationally less expensive as compared to the latter. Therefore, we choose traditional \emph{ML} classification algorithms to verify the impact of various features present in programming codes on their runtime complexities. We also perform a similar analysis on a simple Multi level Perceptron\emph{(MLP)} classifier and compare against others. Table \ref{table:score_precision_recall} depicts the accuracy score, weighted precision and recall values for this classification task using 8 different algorithms, with the best accuracy score achieved using the ensemble approach of random forests. 
    
\begin{table}[h]
    \begin{minipage}{.5\linewidth}
      \centering
        \begin{tabular}{|p{23mm}|p{13mm}|p{13mm}|p{9mm}|}
            \hline
            Algorithm & Accuracy & Precision & Recall\\
            \hline\hline
            K-means & 54.76 & 54.34 & 53.95  \\
            Random forest & \textbf{74.26} & 70.85 & 73.19 \\
            Naive Bayes & 57.75 & 60.35 & 58.06 \\
            k-Nearest & 59.89 & 59.35 & 58.57 \\
            Logistic Regression & 73.19 & 72.89 & 73.19 \\
            Decision Tree & 73.79 & 71.86 & 71.12 \\
            MLP Classifier & 63.10 & 59.13 & 58.28 \\
            \textbf{SVM} & \textbf{72.96} & \textbf{69.43} & \textbf{70.58} \\\hline\hline
            \end{tabular}
            \caption{Accuracy Score, Precision and Recall values for different classification algorithms}
            \label{table:score_precision_recall}
    \end{minipage}%
    \begin{minipage}{.5\linewidth}
      \centering
        \begin{tabular}{|p{32mm}|p{22mm}|}
                \hline
                Feature & Mean Accuracy \\
                \hline\hline
                No. of ifs &  44.35\\
                No. of switches & 44.38\\
                No. of loops & 51.33\\
                No. of breaks & 43.85\\
                Priority Queue present & 45.45\\
                No. of sorts & 52.40\\
                Hash Set present & 44.38\\
                Hash Map present & 43.85\\
                Recursion present & 42.38\\
                Nested loop depth & \textbf{66.31}\\
                No. of Variables & 42.78\\
                No. of methods & 42.19\\
                No. of jumps & 43.65\\
                No. of statements & 44.18\\
                \hline\hline
                \end{tabular}
                \caption{Per feature accuracy score, averaged over different classification algorithms.}
                \label{table:per feature}
    \end{minipage} 
\end{table}


Further, as per Table \ref{table:per feature} showing per-feature-analysis, we distinctly make out that for the collected dataset, the most prominent feature which solely gives maximum accuracy is nested loop depth, followed by number of sorts and loops. For all algorithms, accuracy peaks uptil the tenth feature, and decreases thereafter \cite{additional_results}. This comprehensibly depicts that for the five complexity classes considered here, the number of variables, number of methods, number of jumps and number of statements do not play a notable role as compared to the other features.

Another interesting feature that comes up with the qualitative study of these accuracy scores is that with these features, most of the true classification predictions come up with code samples of category $O{(1)}$, $O{(n)}$ and $O{(n}^2)$. This holds a subtle resonance with the inherent obvious instincts of a human classifier who is more likely to correctly classify a program for these complexities instantly as compared to $O{(logn)}$ or $O{(nlogn)}$. Tables \ref{table:accuracy_3_classes} and \ref{table:accuracy_3_classes_2} demarcate the difference between accuracy scores considering data samples from classes $O{(1)}$, $O{(n)}$, $O{(n}^2)$ as compared to classes $O{(1)}$, $O{(logn)}$, $O{(nlogn})$. A clear increment in accuracy scores is noticed amongst all the algorithms considered for the classification task for the former set of 3 classes considered as compared to the latter.


\begin{table}[!htb]
    \begin{minipage}{.5\linewidth}
      \centering
        \begin{tabular}{|p{20mm}|p{13mm}|p{13mm}|p{9mm}|}
        \hline
        Algorithm & Accuracy & Precision & Recall\\
        \hline\hline
        K-means & 64.38 & 63.76 & 63.04 \\
        Random forest & \textbf{82.61} & 81.32 & 79.37 \\
        Naive Bayes & 63.42 & 61.44 & 61.09 \\
        k-Nearest & 65.56 & 67.54 & 66.23 \\
        Logistic Regression & 81.58 & 81.70 & 80.98 \\
        Decision Tree & 80.12 & 81.19 & 78.24 \\
        MLP Classifier & 70.06 & 65.53 & 70.31 \\
        SVM & 76.43 & 72.14 & 74.35 \\\hline\hline 
        \end{tabular}
        \caption{Accuracy, Precision and Recall values for different classification algorithms considering samples from complexity classes $O{(1)}$, $O{(n)}$ and $O{(n}^2)$}
        \label{table:accuracy_3_classes}
    \end{minipage}%
    \begin{minipage}{.5\linewidth}
      \centering
        \begin{tabular}{|p{20mm}|p{13mm}|p{13mm}|p{9mm}|}
        \hline
        Algorithm & Accuracy & Precision & Recall\\
        \hline\hline
        K-means & 52.31 & 53.23 & 50.04 \\
        Random forest & 64.05 & 71.21 & 63.24 \\
        Naive Bayes & 61.05 & 68.21 & 60.24 \\
        k-Nearest & 63.56 & 64.45 & 64.07 \\
        Logistic Regression & 75.46 & 71.24 & 70.98 \\
        Decision Tree & 74.74 & 72.12 & 77.05 \\
        MLP Classifier & 67.34 & 68.45 & 67.43 \\
        SVM & 69.64 & 70.76 & 67.24 \\\hline\hline
        \end{tabular}
\caption{Accuracy, Precision and Recall values for different classification algorithms considering samples from complexity classes $O{(1)}$, $O{(logn)}$ and $O{(nlogn})$}
\label{table:accuracy_3_classes_2}
    \end{minipage} 
\end{table}

\subsection {Code Embeddings}
We extracted ASTs from source codes, computed 1024-dimensional code embeddings from ASTs using graph2vec 
and trained an SVM classifier on these embeddings. Results are tabulated in Tables \ref{table:graph2vec_results}.
We note that the maximum accuracy obtained for SVM on code embeddings is comparable to that of SVM on statistical features. Also, code embeddings baseline has better precision and recall scores for both representations of AST.

\subsection{Data Ablation Experiments}
To get further insight into the learning framework, we perform following data ablation tests:

\emph{Label Shuffling.} Training models with shuffled class labels can indicate whether the model is learning useful features pertaining to the task at hand. If the performance does not significantly decrease upon shuffling, it can imply that the model is hanging on to statistical cues that do not contain meaningful information w.r.t. the problem.

\emph{Method/Variable Name Alteration.} Graph2vec uses node labels along with edge information to generate graph embeddings. Out of randomly selected 50 codes having correct prediction,
if the predicted class labels before and after data ablation are different for a significant number of test samples, it would imply
it would imply that the model relies on method/variable name tokens whereas it should only rely on the relationships between variables/methods.

\emph{Replacing Input Variables with Constant Literals.} Program complexity is a function of input variables. Thus, to test the robustness of models, we replace the input variables with constant values making resultant complexity $O(1)$ for 50 randomly chosen codes, which earlier had non-constant 
complexity. A good model should have a higher percentage of codes with predicted complexity as  $O(1)$. 

\emph{Removing Graph Substructures. }We randomly remove program elements such as \emph{for}, \emph{if} blocks with a probability of 0.1. The expectation is that the correctly predicted class labels should not change heavily as the complexity most likely does not change and hence should have a higher percentage of codes with same correct label before and after removing graph substructures. 
This would imply that the model is robust to changes in code that do not change the resultant complexity.

\begin{table}[h]
{\small 
\centering
	{\fontsize{8}{9}\selectfont
\begin{tabular}{|p{50mm}|p{15mm}|p{15mm}|p{12mm}|}
\hline
AST Representation & Accuracy & Precision & Recall\\
\hline\hline
Node Labels with concatenation & 73.86 & 74 & 73 \\
Node Labels without concatenation & 70.45 & 71 & 70 \\
\hline\hline 
\end{tabular}
}
\caption{Accuracy, Precision, Recall values for classification of graph2vec embeddings, with and without node type \& node value concatenation in node label.}
\label{table:graph2vec_results}
}
\end{table}

Following are our observations regarding data ablation results in Table \ref{table:graph2vec_results2}:

\emph{Label Shuffling.} The drop in test performance is higher in graph2vec than that in the basic model indicating that graph2vec learns better features compared to simple statistical models.

\emph{Method/Variable Name Alteration.} Table \ref{table:graph2vec_results2} shows that SVM correctly classifies most of the test samples' embeddings upon altering method and variable names, implying that the embeddings generated do not rely heavily on the actual method/variable name tokens.

\emph{Replacing Input Variables with Constant Literals.} We see a significant and unexpected dip in accuracy, highlighting one of the limitations of our model.  

\emph{Removing Graph Substructures.} Higher accuracy for code embeddings as compared to feature engineering implies that the model must be learning the types of nodes and their effect on complexity to at least some extent, as removing substructures does not change the predicted complexity class of a program significantly.
\begin{table}[h]
{\small 
\centering
	{\fontsize{8}{9}\selectfont
\begin{tabular}{|p{44mm}|p{16mm}|p{24mm}|p{24mm}|}
\hline
\multirow{2}{*}{Ablation Technique} 
& \multicolumn{3}{|c|}{Accuracy} \\
\cline{2-4} & Feature Engineering & Graph2vec: With Concatenation & Graph2vec: Without Concatenation\\
\hline\hline
Label Shuffling & 48.29 & 36.78 & 31.03 \\ 
        Method/Variable Name Alteration & NA & 84.21 & 89.18 \\
        Replacing Input Variables with Constant Literals & NA & 16.66 & 13.33 \\
        Removing Graph Substructures & 66.92 & 87.56 & 88.96 \\
\hline\hline 
\end{tabular}
}
\caption{Data Ablation Tests Accuracy of feature engineering and code embeddings \emph{(for two different AST representations)} baselines}
\label{table:graph2vec_results2}
}
\end{table}
\section{Limitations}
The most pertinent limitation of the dataset is its size which is fairly small compared to what is considered standard today. 
Another limitation of our work is low accuracy of the models. An important point to note is that although we established that using code embeddings is a better approach, still their accuracy does not beat feature engineering significantly. One possible solution is to increase dataset size so that generated code embeddings can better model the characteristics of programs that differentiate them into multiple complexity classes, when trained on larger number of codes.
However, generating a larger dataset is a challenging task since annotation process is tedious and needs people with a sound knowledge of algorithms. Lastly, we observe that replacing variables with constant literals does not change the prediction to $O(1)$ which highlights the inability of graph2vec to identify the variable on which complexity depends.
\section{Usefulness of the Dataset}
Computational complexity is a quantification of computational efficiency. Computationally efficient programs better utilize resources and improve software performance. With rapid advancements, there is a growing demand for resources; at the same time, there is greater need for optimizing existing solutions. Thus, writing computationally efficient programs is an asset for both students and professionals. With this dataset, we aim to analyze attributes and capture relationships that best define the computational complexity of codes. We do so, not just by heuristically picking up evident features, but by investigating their role in the quality, structure and dynamics of the problem using \emph{ML} paradigm. We also capture relationships between various programming constructs by generating code embeddings from Abstract Syntax Trees.
This dataset can not only help automate the process of predicting complexities, but also, to help learners decide apt features for well-structured and efficient codes. It is essential for better programming ethics apart from a good algorithm, since a good algorithm might get restricted performance due to inefficient coding, which  can add an additional ease and value to educational purposes.
It can also be used to train models that can be further integrated with IDEs and assist professional developers in writing computationally efficient programs for fast performance software development.
\section{Conclusion}
The dataset presented  and the baseline models established should serve as guidelines for the future work in this area. The dataset presented is balanced and well-curated. Though both the baselines; Code Embeddings and Handcrafted features have comparable accuracy, we have established through data ablation tests that code embeddings learned from Abstract Syntax Tree of the code better capture relationships between different code constructs that are essential for predicting runtime complexity. Work can be done in future to increase the size of the dataset to verify our hypothesis that code embeddings will perform significantly better than hand crafted features. Moreover, we hope that the approaches discussed in this work, their usage becomes explicit for programmers and learners to bring into practice efficient and optimized codes. 
%
%
%
%
\bibliographystyle{splncs04}
\bibliography{references}

\begin{thebibliography}{10}
\providecommand{\url}[1]{\texttt{#1}}
\providecommand{\urlprefix}{URL }
\providecommand{\doi}[1]{https://doi.org/#1}

\bibitem{haltingproof}
Are runtime bounds in p decidable? (answer: no)
  \url{https://cstheory.stackexchange.com/questions/5004/are-runtime-bounds-in-p-decidable-answer-no}

\bibitem{graph2vec_implementation}
Graph2vec implementation \url{https://github.com/MLDroid/graph2vec_tf}

\bibitem{additional_results}
More additional results at
  \url{https://github.com/midas-research/corcod-dataset}

\bibitem{pmlr-v48-allamanis16}
Allamanis, M., Peng, H., Sutton, C.: A convolutional attention network for
  extreme summarization of source code. In: Balcan, M.F., Weinberger, K.Q.
  (eds.) Proceedings of The 33rd International Conference on Machine Learning.
  Proceedings of Machine Learning Research, vol.~48, pp. 2091--2100. PMLR, New
  York, New York, USA (20--22 Jun 2016),
  \url{http://proceedings.mlr.press/v48/allamanis16.html}

\bibitem{ASTPaths-code2vec}
Alon, U., Zilberstein, M., Levy, O., Yahav, E.: A general path-based
  representation for predicting program properties. CoRR
  \textbf{abs/1803.09544} (2018), \url{http://arxiv.org/abs/1803.09544}

\bibitem{code2vec}
Alon, U., Zilberstein, M., Levy, O., Yahav, E.: Code2vec: Learning distributed
  representations of code. Proc. ACM Program. Lang.  \textbf{3}(POPL),
  40:1--40:29 (Jan 2019). \doi{10.1145/3290353},
  \url{http://doi.acm.org/10.1145/3290353}

\bibitem{ricetheorem}
Asperti, A.: The intensional content of rice's theorem. In: Proceedings of the
  35th Annual ACM SIGPLAN-SIGACT Symposium on Principles of Programming
  Languages. pp. 113--119. POPL '08, ACM, New York, NY, USA (2008).
  \doi{10.1145/1328438.1328455},
  \url{http://doi.acm.org/10.1145/1328438.1328455}

\bibitem{master}
Bentley, J.L., Haken, D., Saxe, J.B.: A general method for solving
  divide-and-conquer recurrences. SIGACT News  \textbf{12}(3),  36--44 (Sep
  1980). \doi{10.1145/1008861.1008865},
  \url{http://doi.acm.org/10.1145/1008861.1008865}

\bibitem{CodeEmbeddings-LiteratureStudy}
Chen, Z., Monperrus, M.: A literature study of embeddings on source code. CoRR
  \textbf{abs/1904.03061} (2019), \url{http://arxiv.org/abs/1904.03061}

\bibitem{doc2vec}
Le, Q.V., Mikolov, T.: Distributed representations of sentences and documents
  (2014)

\bibitem{DefectPred}
Li, J., He, P., Zhu, J., Lyu, M.R.: Software defect prediction via
  convolutional neural network. 2017 IEEE International Conference on Software
  Quality, Reliability and Security (QRS) pp. 318--328 (2017)

\bibitem{public_git_archive_dataset}
Markovtsev, V., Long, W.: Public git archive: a big code dataset for all. CoRR
  \textbf{abs/1803.10144} (2018), \url{http://arxiv.org/abs/1803.10144}

\bibitem{mccabe}
McCabe, T.J.: A complexity measure. In: Proceedings of the 2Nd International
  Conference on Software Engineering. pp. 407--. ICSE '76, IEEE Computer
  Society Press, Los Alamitos, CA, USA (1976),
  \url{http://dl.acm.org/citation.cfm?id=800253.807712}

\bibitem{graph2vec}
Narayanan, A., Chandramohan, M., Venkatesan, R., Chen, L., Liu, Y., Jaiswal,
  S.: graph2vec: Learning distributed representations of graphs. CoRR
  \textbf{abs/1707.05005} (2017), \url{http://arxiv.org/abs/1707.05005}

\bibitem{AspiringMinds}
Srikant, S., Aggarwal, V.: A system to grade computer programming skills using
  machine learning. In: Proceedings of the 20th ACM SIGKDD International
  Conference on Knowledge Discovery and Data Mining. pp. 1887--1896. KDD '14,
  ACM, New York, NY, USA (2014). \doi{10.1145/2623330.2623377},
  \url{http://doi.acm.org/10.1145/2623330.2623377}

\bibitem{Sun_Zhu_Mou_Xiong_Li_Zhang_2019}
Sun, Z., Zhu, Q., Mou, L., Xiong, Y., Li, G., Zhang, L.: A grammar-based
  structural cnn decoder for code generation. Proceedings of the AAAI
  Conference on Artificial Intelligence  \textbf{33}(01),  7055--7062 (Jul
  2019). \doi{10.1609/aaai.v33i01.33017055},
  \url{https://www.aaai.org/ojs/index.php/AAAI/article/view/4686}

\bibitem{StacQ}
Yao, Z., Weld, D.S., Chen, W., Sun, H.: Staqc: {A} systematically mined
  question-code dataset from stack overflow. CoRR  \textbf{abs/1803.09371}
  (2018), \url{http://arxiv.org/abs/1803.09371}

\bibitem{so_dataset}
Yin, P., Deng, B., Chen, E., Vasilescu, B., Neubig, G.: Learning to mine
  aligned code and natural language pairs from stack overflow. In:
  International Conference on Mining Software Repositories. pp. 476--486. MSR,
  ACM (2018). \doi{https://doi.org/10.1145/3196398.3196408}

\bibitem{MethodNamePred-CallGraphEmbedding}
Yonai, H., Hayase, Y., Kitagawa, H.: Mercem: Method name recommendation based
  on call graph embedding. CoRR  \textbf{abs/1907.05690} (2019),
  \url{http://arxiv.org/abs/1907.05690}

\end{thebibliography}
\end{document}